\newif\ifsingle
\singletrue 

\newif\ifproofs

\ifsingle
\documentclass[12pt,draftclsnofoot, onecolumn]{IEEEtran}		
\else		
\documentclass[10pt,final, twocolumn]{IEEEtran}
\fi

\usepackage{times}
\usepackage{amsmath,dsfont}
\usepackage{amssymb,amsthm}
\usepackage{epsfig,verbatim}
\usepackage{subfigure}
\usepackage{setspace}
\usepackage{color}
\usepackage{cite}
\usepackage{epstopdf}
\usepackage{graphics}
\usepackage{accents}
\usepackage{acronym}
\usepackage[bookmarks,colorlinks]{hyperref}
\usepackage{booktabs}
\usepackage{mathtools}
\usepackage{algorithm}
\usepackage{algorithmic}
\usepackage{enumitem}

\newtheorem{example}{Example}

\newcommand{\myVec}[1]{{\boldsymbol{#1}}}

\newcommand{\mySet}[1]{\mathcal{#1}}
\newcommand{\myX}{{\myVec{x}}}			 		

\acrodef{dnn}[DNN]{deep neural network} 
\acrodef{csi}[CSI]{channel state information}
\acrodef{map}[MAP]{maximum a-posteriori probability}
\acrodef{snr}[SNR]{signal-to-noise ratio}
\acrodef{ser}[SER]{symbol error rate}
\acrodef{bs}[BS]{base station} 
\acrodef{ml}[ML]{machine learning} 
\acrodef{iot}[IOT]{Interent of Things}
\acrodef{mimo}[MIMO]{multiple-input multiple-output}
\acrodef{mse}[MSE]{mean-squared error}
\acrodef{pdf}[PDF]{probability density function}
\acrodef{rv}[RV]{random variable}
\acrodef{em}[EM]{expectation maximization}
\acrodef{hmm}[HMM]{hidden Markov model}
\acrodef{pdf}[PDF]{probability density function}
\acrodef{isi}[ISI]{intersymbol interference}  
\acrodef{awgn}[AWGN]{additive white Gaussian noise}

\ifsingle

\newcommand{\includefig}[1]{\includegraphics[width = 0.75\columnwidth]{#1} 	\vspace{-0.2cm}}
\else

\newcommand{\includefig}[1]{\includegraphics[width = \columnwidth]{#1} 	\vspace{-0.2cm}}

\fi 

\IEEEoverridecommandlockouts

\title{Data-Driven Factor Graphs for Deep Symbol Detection
}
\author{
	\IEEEauthorblockN{Nir Shlezinger,  Nariman Farsad, Yonina C. Eldar, and Andrea J. Goldsmith\\
	} 
	\thanks{This work was supported in part by the  US - Israel Binational Science Foundation	under grant No. 2026094,  by the Israel Science Foundation under grant No. 0100101, and by the Office of the Naval Research under grant No. 18-1-2191.
		N. Shlezinger  and Y. C. Eldar are with the Faculty of Math and CS, Weizmann Institute of Science, Rehovot, Israel (e-mail: nirshlezinger1@gmail.com; yonina@weizmann.ac.il). 	
		N. Farsad  and A. J. Goldsmith are with the Department of EE, Stanford, Palo Alto, CA (e-mail:  nfarsad@stanford.edu; andrea@wsl.stanford.edu).  	
	}

	\vspace{-1.0cm}
	
}
\vspace{-0.75cm}

\begin{document}

\maketitle
\begin{abstract}
	Many important schemes in signal processing and communications, ranging from the BCJR algorithm to the Kalman filter, are instances of factor graph methods. This family of algorithms is based on recursive message passing-based computations carried out over graphical models, representing a factorization of the underlying statistics. Consequently, in order to implement these algorithms, one must have accurate knowledge of the statistical model of the considered signals. In this work we propose to implement factor graph methods in a data-driven manner. In particular, we propose to use machine learning (ML) tools to learn the factor graph, instead of the overall system task, which in turn is used for inference by message passing over the learned graph. We apply the proposed approach to learn the factor graph representing a finite-memory channel, demonstrating the resulting ability to  implement BCJR detection in a data-driven fashion. We demonstrate that the proposed system, referred to as BCJRNet, learns to implement the BCJR algorithm from a small training set, and that the resulting receiver exhibits improved robustness to inaccurate training compared to the conventional channel-model-based receiver operating under the same level of uncertainty. Our results indicate that by utilizing ML tools to learn factor graphs from labeled data, one can implement a broad range of model-based algorithms, which traditionally require full knowledge of the underlying statistics, in a data-driven fashion. 
\end{abstract}
\vspace{-0.4cm}
\section{Introduction}
\vspace{-0.1cm}
A broad range of algorithms in communications, signal processing, statistics, and machine learning are obtained by message passing over graphical models \cite{loeliger2007factor}. In particular, the combination of factor graphs, and specifically, the factor graph model proposed by Forney in \cite{forney2001codes}, with the sum-product message passing scheme \cite{kschischang2001factor}, was shown to provide a unified framework which specializes many important algorithms~\cite{loeliger2004introduction}.  

Broadly speaking, factor graphs provide a graphical model for a function of multiple variables, commonly a joint distribution measure. 
The resulting model allows prohibitive computations to be carried out in a recursive manner  at controllable complexity by message passing along the graph~\cite{loeliger2004introduction}. Some important instances of factor graph methods include the BCJR symbol detector \cite{bahl1974optimal};  the Kalman filter \cite{loeliger2002least};  and \ac{hmm} predictors~\cite{jordan2004graphical}. Furthermore, the application of message passing over factor graphs was shown to give rise to efficient and accurate methods for, e.g., equalization \cite{colavolpe2005application, drost2007factor}, \acl{mimo} detection \cite{som2011low}, and  joint decoding and channel estimation \cite{niu2005factor, lehmann2016factor}.

The aforementioned algorithms require prior knowledge on the underlying statistical model. For example, in order to apply the BCJR symbol detector, one must know or have an accurate estimate of the channel model and its parameters, i.e., full \ac{csi}. This limits the application of these methods in setups where the statistical model is complex, difficult to estimate, or poorly understood. Furthermore, a potentially large overhead is required in order to estimate the model parameters, from which the factor graph is obtained, and the reliability of message-passing methods may be substantially degraded in the presence of model inaccuracy. 

In our previous works \cite{shlezinger2019viterbinet} and \cite{shlezinger2019deepSIC} we proposed data-driven receivers which learn to implement two important symbol detection algorithms: the Viterbi algorithm \cite{forney1973viterbi}, and the iterative soft interference cancellation scheme \cite{choi2000iterative}. By using dedicated \ac{ml} tools to learn the graphical model used by the channel-model-based symbol detector algorithm, e.g., the trellis diagram for the  Viterbi algorithm, we obtained systems which implement the symbol detection scheme in a data-driven fashion using a relatively small amount of training samples, while exhibiting improved robustness to \ac{csi} uncertainty compared to their model-based counterpart. Our success in implementing a model-based algorithm in a data-driven manner by learning its underlying graphical model, combined with fact that the Viterbi algorithm is a factor graph method, motivates the study of a data-driven implementation for a broader family of factor graph methods.

In this paper we take a first step in examining the ability to implement data-driven algorithms by utilizing \ac{ml} tools to learn the underlying factor graph. We focus on problems which can be solved using factor graph methods when the model is known, proposing to utilize \ac{ml} tools to learn the factor graph, and to carry out the task using conventional message passing over the learned graph. This is opposed to the common approach in \ac{ml}, which is based on training \acp{dnn} to carry out the task in an end-to-end manner \cite{lecun2015deep}. This approach which combines \ac{ml} and model-based algorithms in a hybrid manner, having the potential of achieving the performance guarantees and controllable complexity of model-based methods while operating in a data-driven manner and requiring relatively small training sets. 

We show how the proposed approach of learning factor graphs can be applied to symbol detection in finite-memory channels. In particular, we detail how this strategy can be used to implement the BCJR algorithm, which is the \ac{map} symbol detector for such channels, in a data-driven channel-model-independent manner. The resulting data-driven receiver, referred to as {\em BCJRNet}, is   capable of learning to carry out \ac{map} symbol detection from a relatively small training set, without requiring prior knowledge of the channel model and its parameters. Our numerical studies also show that the BCJRNet exhibits improved resiliency to inaccurate training compared to the model-based BCJR algorithm operating under the same level of uncertainty. Our results demonstrate the potential gains of properly combining \ac{ml} and model-based algorithms in realizing optimal-approaching data-driven methods in communications and signal processing.  

The rest of this paper is organized as follows: In Section~\ref{sec:FG} we briefly review factor graph methods, and propose the concept of data-driven factor graphs. Section~\ref{sec:SymbolDet}  presents the application of this approach to realize a data-driven BCJR detector. Section~\ref{sec:Sims} provides numerical performance results of BCJRNet, while Section~\ref{sec:Conclusions} provides concluding remarks. 

Throughout the paper, we use upper-case letters for \acp{rv}, e.g. $X$.
Boldface lower-case letters denote vectors, e.g., ${\myVec{x}}$ is a deterministic vector, and $\myVec{X}$ is a random vector;
the $i$th element of ${\myVec{x}}$ is written as $\left( {\myVec{x}}\right) _i$. 
The probability measure of an \ac{rv} $X$ evaluated at $x$ is denoted $P_X(x)$. We use caligraphic letters for sets, e.g., $\mySet{X}$, where $|\mySet{X}|$ is the cardinality of a finite set $\mySet{X}$, and 
$\mySet{R}$ is the set of real numbers.

\vspace{-0.2cm}
\section{Data-Driven Factor Graphs}
\label{sec:FG}
\vspace{-0.1cm}
In this section we present the concept of data-driven factor graphs. We begin with a brief review of conventional model-based factor graph methods in Subsection \ref{subsec:FGModel}, after which we discuss how these graphs can be learned from labeled data using \ac{ml} tools in Subsection \ref{subsec:FGLearn}.

\vspace{-0.2cm}
\subsection{Model-Based Factor Graphs}
\label{subsec:FGModel}
\vspace{-0.1cm}
In the following we provide a brief introduction to factor graphs, focusing on the model proposed by Forney in \cite{forney2001codes}, known as {\em Forney-style factor graphs}. We then review the sum-product method for computing marginal distributions using factor graphs \cite{kschischang2001factor}. The main example considered in the sequel is that of the factor graph representation of finite-state channels, which is utilized in Section \ref{sec:SymbolDet} for formulating the data-driven BCJRNet \ac{map} symbol detector. 

A factor graph is a graphical representation of the factorization of a function of several variables \cite{loeliger2004introduction}, commonly a joint distribution measure. Consider an $n \times 1$ random vector $\myVec{X} \in \mySet{X}^n$ where $\mySet{X}$ is a finite set, i.e., the entries of $\myVec{X}$, denoted $\{X_i\}$, are discrete \acp{rv}. The joint distribution of $\myVec{X}$, $P_{\myVec{X}}(\myX)$, can be factorized if it can be represented as the product of $m$ functions $\{f_i(\cdot)\}_{i=1}^{m}$, i.e., there exists some partition variables $\{\mySet{V}_i\}_{i=1}^{m}$ where $\mySet{V}_i \subset \{x_1, \ldots, x_n\}$ such that 
\begin{equation}
P_{\myVec{X}}(\myX) = \prod_{i=1}^{m} f_i(\mySet{V}_i).
\label{eqn:Decomp}
\end{equation} 
In order to represent \eqref{eqn:Decomp} using a factor graph, the functions $\{f_i(\cdot)\}_{i=1}^{m}$ should be set such that each variable $x_k$ appears in no more than two partitions\footnote{A factorization in which a variable appears in more than two factors can always be modified to meet the above constraint by introducing additional variables and identity factors, see \cite{loeliger2004introduction}.}  $\{\mySet{V}_i\}_{i=1}^{m}$. Subject to this assumption, the joint distribution $P_{\myVec{X}}(\myX)$ can be represented using a factor graph with $m$ factor nodes, which are the functions  $\{f_i(\cdot)\}_{i=1}^{m}$, and $n$ edges $\{x_k\}_{k=1}^{n}$, where variables appearing only in a single partition are treated as half-edges.

\begin{figure}
	\centering
	{\includefig{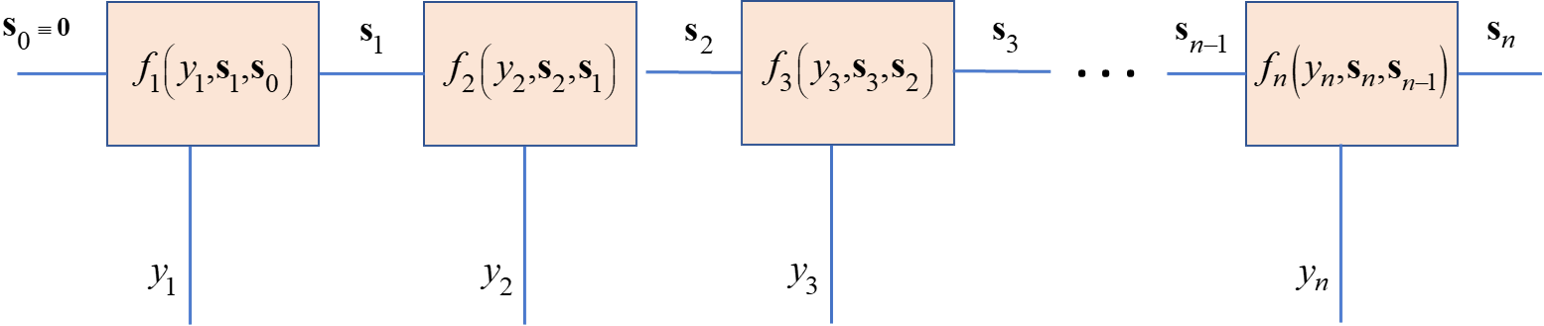}} 
	\caption{Factor graph of finite-memory channel.}
	\label{fig:FGFinite}	 
\end{figure}

\begin{example}
	\label{exm:FSChannel}
	A {\em finite-memory channel} models a statistical relationship between a channel input $X_i \in \mySet{X}$ and a  channel output $Y_i$, such that $Y_i$ depends only on the last $L>0$ channel inputs $X_i,\ldots X_{i-L+1}$. For a  block of $n$ i.i.d. input symbols  $\myVec{X}$, the channel input-output relationship satisfies 
	\begin{equation}
	P_{\myVec{Y} |\myVec{X}}(\myVec{y}|\myVec{x}) = \prod_{i=1}^{n}P_{Y_i|X_i,\ldots,X_{i=L+1}}(y_i|x_i,\ldots,x_{i-L+1}),
	\end{equation}
	where $X_i \equiv 0$ for $i \le 0$. By defining the $L \times 1$ state vector $\myVec{S}_i \triangleq [X_i,\ldots X_{i-L+1}]^T$ and stacking the states over the entire block into a vector  $\myVec{S} \in \mySet{X}^{Ln}$, we can write the joint state-output distribution as
	\begin{align}
P_{\myVec{Y}, \myVec{S}}(\myVec{y}, \myVec{s} )  
	&	= P_{\myVec{Y}| \myVec{S}}(\myVec{y}| \myVec{s} )  P_{\myVec{S}}(\myVec{s} )  \notag \\
	&= \prod_{i=1}^{n}	P_{Y_i| \myVec{S}_i}\left( y_i| \myVec{s}_i\right)	P_{\myVec{S}_i| \myVec{S}_{i-1}}\left( \myVec{s}_i| \myVec{s}_{i-1}\right). 
	\label{eqn:FSC2} 
	\end{align}
%
%
	By defining the functions 
	\begin{equation}
	\label{eqn:FSC_funcNode}
	f_i(y_i, \myVec{s}_i, \myVec{s}_{i-1}) \triangleq P_{Y_i| \myVec{S}_i}\left( y_i| \myVec{s}_i\right)	P_{\myVec{S}_i| \myVec{S}_{i-1}}\left( \myVec{s}_i| \myVec{s}_{i-1}\right),
	\end{equation}
	  for each $i\in\{1,\ldots,n\} \triangleq  \mySet{N}$, it holds that the channel input-output relationship \eqref{eqn:FSC2} obeys the form of \eqref{eqn:Decomp}. The resulting factor graph, in which  $\{\myVec{s}_i\}_{i=1}^{n-1}$ are the edges while the remaining variables are half-edges, is depicted in Fig. \ref{fig:FGFinite}.
\end{example}

\begin{figure}
\centering
{\includefig{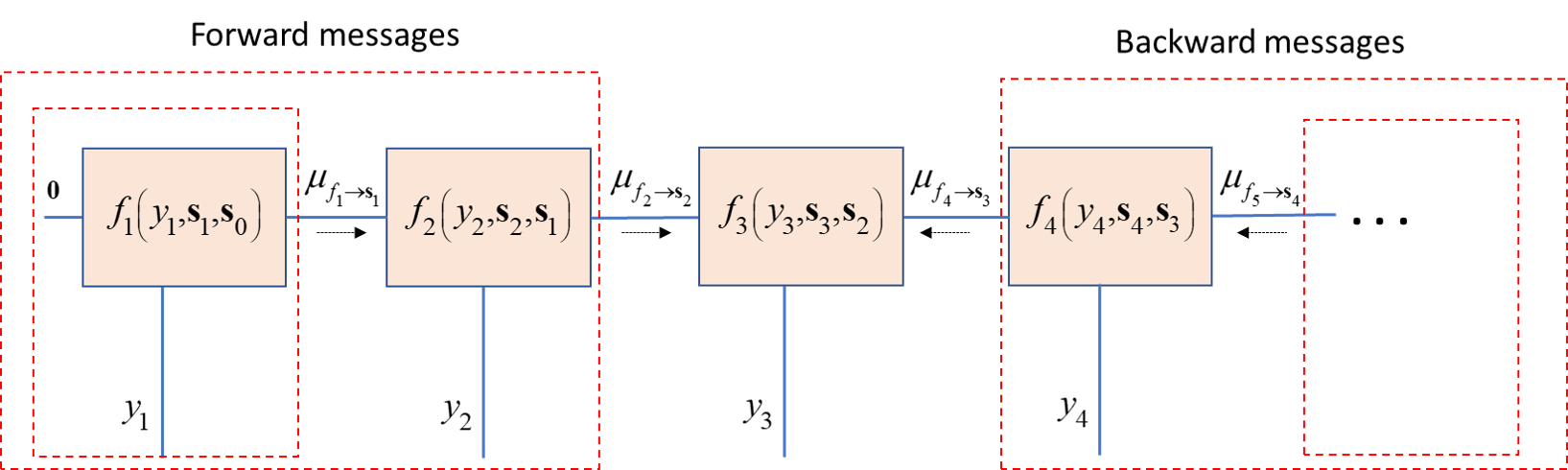}} 
\caption{Message passing over the factor graph of a finite-memory channel.}
\label{fig:SumProduct2}	 
\end{figure}

A major advantage for representing joint distributions via factor graphs is that they allow some  desired computations to be carried out with reduced complexity. One of the most common methods is the sum-product algorithm for evaluating a marginal distribution from a factor graph representation of a joint probability measure \cite{kschischang2001factor}.  To formulate this method, consider the factorized distribution \eqref{eqn:Decomp}, where the ordering of the partitions $\{\mySet{V}_i\}$ corresponds to the order of the variables $\{x_k\}$, e.g., $\mySet{V}_1 = \{x_1, x_2\}$, $\mySet{V}_2 = \{x_2, x_3, x_4\}$, $\mySet{V}_3 = \{x_4, x_5\}$, etc. Furthermore, assume that the factor graph of \eqref{eqn:Decomp} does not contain any cycles\footnote{While there  the sum-product algorithm  can be extended for handling cycles, e.g., \cite{weiss2001optimality}, we focus here on the standard application for cycle-free graphs \cite{loeliger2004introduction,loeliger2007factor}.}. In this case, the marginal distribution of a single \ac{rv} $X_k$ whose corresponding variable appears in the partitions $\mySet{V}_{j}$ and $\mySet{V}_{j+1}$ can be computed from the joint distribution of $\myVec{X}$ via 
\begin{align}
&P_{X_k}(x_k) 
= \sum_{\{x_1,\ldots,x_{k-1}, x_{k+1},\ldots, x_n\}}  P_{\myVec{X}}(x_1,\ldots, x_n) \notag \\
&= \underbrace{\left(\sum_{\{x_1 \ldots x_{k\!-\!1}\}} \prod_{i=1}^{j} f_i(\mySet{V}_i) \right)}_{\triangleq \mu_{f_{j} \rightarrow x_k}( x_k)} \underbrace{\left(\sum_{\{x_{k\!+\!1}\ldots x_{n}\}} \prod_{i={j\!+\!1}}^{m} f_i(\mySet{V}_i) \right)}_{\triangleq \mu_{f_{j\!+\!1} \rightarrow x_k}( x_k)}.
\label{eqn:MarginalFG}
\end{align}
The factorization of the joint distribution in \eqref{eqn:Decomp} implies that the marginal distribution, whose computation typically requires summation over $|\mySet{X}|^{n-1}$ variables, can now be evaluated as the product of two terms, $\mu_{f_{j} \rightarrow x_k}( x_k)$ and   $ \mu_{f_{j+1} \rightarrow x_k}( x_k)$. These terms can be viewed as messages propagating forward and backward along the factor graph, e.g., $\mu_{f_{j} \rightarrow x_k}( x_k)$ represents a forward message conveyed from function node $f_{j}$ to edge $x_k$. In particular, these messages can be computed recursively, e.g., by writing $\mySet{V}_j / x_k = \{x_{k-l}, \ldots, x_{k-1}\}$ for some $l\ge 1$, then the sum-product rule \cite{loeliger2004introduction} implies that 
\begin{align}
\mu_{f_{j} \rightarrow x_k}( x_k) &= \sum_{\{x_1,\ldots,x_{k-1}\}} \prod_{i=1}^{j} f_i(\mySet{V}_i) \notag \\
&= \sum_{\{x_{k-l},\ldots,x_{k-1}\}}   f_j(\mySet{V}_j) \prod_{i=k-l}^{k-1} \mu_{\tilde{f}_{i,j} \rightarrow x_{i}}( x_{i}),
\end{align} 
where $\tilde{f}_{i,j}$ is the function node connected to the edge $x_i$ other than $f_j$.
This method of computing marginal distributions using forward and backward recursions along the factor graph is referred to as belief propagation \cite{pearl1986fusion}, or the sum-product algorithm \cite{kschischang2001factor}. The application of this scheme in finite-memory channels  is detailed in the following example:
\begin{example}
	\label{exm:SP_FSC}
	Consider again the setup discussed in Example \ref{exm:FSChannel} representing a channel with finite memory of length $L>0$. We are interested in computing the joint distribution of two consecutive state vectors, $\myVec{S}_k$ and $\myVec{S}_{k+1}$, given a realization of the channel output $\myVec{Y} = \myVec{y}$. Using the sum-product method, one can compute this joint distribution by recursive message passing along its factor graph. In particular, 
	\begin{align}
	P_{\myVec{S}_k,\myVec{S}_{k+1}, \myVec{Y}}(\myVec{s}_k,\myVec{s}_{k+1}, \myVec{y}) &= \mu_{f_k \rightarrow \myVec{s}_k}(\myVec{s}_k) f_{k+1}(y_{k+1},\myVec{s}_{k+1}, \myVec{s}_k) \notag \\
	&\times \mu_{f_{k+2} \rightarrow \myVec{s}_{k+1}}(\myVec{s}_{k+1}),
	\label{eqn:Recursion1}
	\end{align}
	where the forward path messages are computed recursively via
	\begin{equation}
	\mu_{f_i \rightarrow \myVec{s}_i}(\myVec{s}_i) = \sum_{\myVec{s}_{i-1}} f_{i}(y_{i},\myVec{s}_{i}, \myVec{s}_{i-1})\mu_{f_{i-1} \rightarrow \myVec{s}_{i-1}}(\myVec{s}_{i-1}),
	\label{eqn:Recursion1Forwards}
	\end{equation}
	for $i = 1, 2,\ldots, k$. Similarly, the backward messages are 
		\begin{equation}
	\mu_{f_{i\!+\!1} \rightarrow \myVec{s}_i}(\myVec{s}_i) = \sum_{\myVec{s}_{i\!+\!1}} f_{i\!+\!1}(y_{i\!+\!1},\myVec{s}_{i\!+\!1}, \myVec{s}_{i})\mu_{f_{i\!+\!2} \rightarrow \myVec{s}_{i\!+\!1}}(\myVec{s}_{i\!+\!1}),
	\label{eqn:Recursion1Backwards}
	\end{equation}
	for $i = n-1,  n-2, \ldots, k+1$. An illustration of this message passing along the factor graph for $k=2$ is  depicted in Fig.~\ref{fig:SumProduct2}.
\end{example}

\vspace{-0.2cm}
\subsection{Learning Factor Graphs}
\label{subsec:FGLearn}
\vspace{-0.1cm}
Factor graph methods, such as the sum-product algorithm, exploit the factorization of a joint distribution to efficiently compute a desired quantity. For example, the application of the sum-product algorithm in a finite-memory channel detailed in Example \ref{exm:SP_FSC} exploits its factorization to compute marginal distributions, an operation whose burden typically grows exponentially with th block size, with complexity that only grows linearly with $n$. As we show in Section \ref{sec:SymbolDet}, the instance of the sum-product algorithm discussed in Example \ref{exm:SP_FSC} is exactly the recursive computation carried out in the BCJR symbol detector \cite{bahl1974optimal}. In fact, the sum-product algorithm specializes a multitude of common signal processing techniques, including the Kalman filter and \ac{hmm} prediction \cite{loeliger2004introduction}. 

In order to implement the sum-product scheme, one must be able to specify the factor graph encapsulating the underlying distribution, and in particular, the function nodes $\{f_j\}_{j=1}^{m}$. This implies that, for example, the BCJR detector, which builds upon the recursive computation in Example \ref{exm:SP_FSC}, requires prior knowledge of the channel model and its parameters, i.e., full \ac{csi}. To realize the Kalman filter, one must first know the underlying state-space equations.  In some case, such accurate prior knowledge may not be available or costly to acquire. 

In our previous works \cite{shlezinger2019viterbinet} we utilized \ac{ml} tools to implement  the Viterbi algorithm, which can also be considered as an instance of a factor graph methods \cite{wiberg1995codes}, in a data-driven fashion. We used a similar approach in \cite{shlezinger2019deepSIC} to realize a data-driven implementation of the soft interference cancellation scheme \cite{choi2000iterative}, which is also a message passing-based symbol detection method. This was achieved by utilizing dedicated compact \acp{dnn} to learn the channel-model-based computations of the underlying graphical model, e.g., the log-likelihood values used for constructing the trellis diagram in the Viterbi detector, while maintaining the overall flow of the algorithm. We propose to generalize this concept to realize a more general family of data-driven factor graph methods by learning the mappings carried out at the function nodes  from a relatively small set of labeled data using \ac{ml} tools. By doing so, one can train a system to learn an underlying factor graph, which can be then utilizing for inference using conventional factor graph methods, such as the sum-product algorithm. 

The proposed approach requires prior knowledge of the graph structure, but not its nodes. For example, a finite-memory channel with memory length not larger than $L$ can be represented using the structure in Fig. \ref{fig:FGFinite} while its specific input-output relationship dictate the function nodes. Consequently, in order to learn such a factor graph from samples, one must only learn its function nodes. As these mappings often represent conditional distribution measures, they can be naturally learned using classification networks, e.g., fully-connected \acp{dnn} with softmax output layer and cross-entropy objective, which are known to reliably learn conditional distributions in complex environments \cite{bengio2009learning}. 
Furthermore, in many scenarios of interest, e.g., stationary finite-memory channels and time-invariant state-space models, the mapping implemented by the factor nodes $f_i(\cdot)$ does not depend on the index $i$. In such cases, only a fixed set of mappings whose size does not grow with the dimensionality $n$ has to be tuned in order to learn the complete factor graph. An example of how this concept of learned factor graphs can be applied is presented in the following section.

\vspace{-0.2cm}
\section{Application: Symbol Detection}
\label{sec:SymbolDet}
\vspace{-0.1cm}
In this section we demonstrate how the concept of data-driven factor graphs can be applied for symbol detection in finite-memory channels. We begin with reviewing the BCJR algorithm, which is the model-based application of the sum-product algorithm for symbol detection, in Subsection \ref{subsec:BCJR}. Then, we show how the factor graph can be learned from training in Subsection \ref{subsec:SymbolDetDeep}, and discuss the pros and cons of the resulting architecture in Subsection \ref{subsec:Discussion}.

\vspace{-0.2cm}
\subsection{Sum-Product for Symbol Detection}
\label{subsec:BCJR}
\vspace{-0.1cm}
We next present the application of the sum-product method for symbol detection in finite-memory channels, also known as the BCJR algorithm \cite{bahl1974optimal}. 
Consider a stationary finite-memory channel, namely, a channel obeying the model in Example~\ref{exm:FSChannel} in which the conditional distribution $P_{Y_k| \myVec{S}_k}(y| \myVec{s})$ does not depend on the time index $k \in \mySet{N}$. Symbol detection refers to the task of recovering the i.i.d. transmitted symbols $\{X_k\}_{k=1}^{n}$, each uniformly distributed over the constellation set $\mySet{X}$, from a realization of the channel output $\myVec{Y} = \myVec{y}$. The detector which minimizes the symbol error probability is the \ac{map} rule
\begin{align}
\hat{X}_k &= \mathop{\arg \max}\limits_{x \in \mySet{X}}P_{X_k|\myVec{Y}}(x|\myVec{y}) \notag \\
&= \mathop{\arg \max}\limits_{x \in \mySet{X}}P_{X_k,\myVec{Y}}(x,\myVec{y}), \qquad k \in \mySet{N}.
\label{eqn:MAP}
\end{align} 
Using the formulation of the state vectors $\{\myVec{S}_k\}$, the desired joint probability can be written as \cite[Ch. 9.3]{cioffi2008equalization}
\begin{equation}
\small
\!\!\!\!P_{X_k,\myVec{Y}}(x,\myVec{y}) \!= \!\!\sum_{\myVec{s}\in\mySet{X}^L} \!\!P_{\myVec{S}_{k\!-\!1},\myVec{S}_{k}, \myVec{Y}}(\myVec{s},[x, (\myVec{s})_1 \ldots  (\myVec{s})_{L\!-\!1}]^T\!, \myVec{y}).
\label{eqn:Joint1}
\end{equation}
The summands in \eqref{eqn:Joint1} are the joint distributions evaluated recursively from the channel factor graph in Example \ref{exm:SP_FSC}. Thus, when the factor graph is known, the \ac{map} rule \eqref{eqn:MAP} can be computed efficiently using the sum-product algorithm.

\vspace{-0.2cm}
\subsection{BCJRNet: Data-Driven MAP Recovery}
\label{subsec:SymbolDetDeep}
\vspace{-0.1cm}
Here, we show how the rationale presented in Section \ref{sec:FG} yields a method for learning the factor graphs of finite-memory channels, using which the \ac{map} detector can be obtained. 
We assume that the channel memory length, $L$, is known. However, the channel model, i.e., the conditional distribution $P_{Y | \myVec{S}}(\cdot)$ is unknown, and only a set of labeled input-output pairs, denoted $\{x_i, y_i\}$ is available.  

Since the channel memory is known, the structure of the factor graph is fixed to that depicted in Fig. \ref{fig:FGFinite}. Consequently, in order to learn the factor graph, one must only adapt the function nodes $f_i(\cdot)$, as discussed in Subsection~\ref{subsec:FGLearn}. Based on \eqref{eqn:FSC2}-\eqref{eqn:FSC_funcNode}, the function nodes 
are given by 
\begin{align}
&f_i(y_i, \myVec{s}_i, \myVec{s}_{i-1}) 
= P_{Y| \myVec{S}}\left( y_i| \myVec{s}_i\right)	P_{\myVec{S}_i| \myVec{S}_{i-1}}\left( \myVec{s}_i| \myVec{s}_{i-1}\right) \notag \\
&\stackrel{(a)}{=} 
\begin{cases}
 \frac{1}{|\mySet{X}|} P_{Y| \myVec{S}}\left( y_i| \myVec{s}_i\right) &(\myVec{s}_i)_j = (\myVec{s}_{i\!-1})_{j\!-\!1}, j=2\ldots L \\
 0 & {\rm otherwise},
\end{cases}
\label{eqn:FunctionNode}
\end{align}
where $(a)$ follows from the definition of the state vectors $\myVec{S}_i$ and the channel stationarity. The formulation of the function nodes in \eqref{eqn:FunctionNode} implies that they can be estimated by training an \ac{ml}-based system to evaluate  $P_{Y | \myVec{S}}(\cdot)$ from which the corresponding function node value is obtained via \eqref{eqn:FunctionNode}. Once the factor graph representing the channel is learned, symbol recovery is carried out using the sum-product method detailed in Subsection \ref{subsec:BCJR}. The resulting receiver, referred to as {\em BCJRNet}, thus implements BCJR detection in a data-driven manner, and is expected to approach \ac{map}-performance when the function nodes are accurately estimated, as demonstrated in our numerical study in Section \ref{sec:Sims}.

\begin{figure}
	\centering
	{\includefig{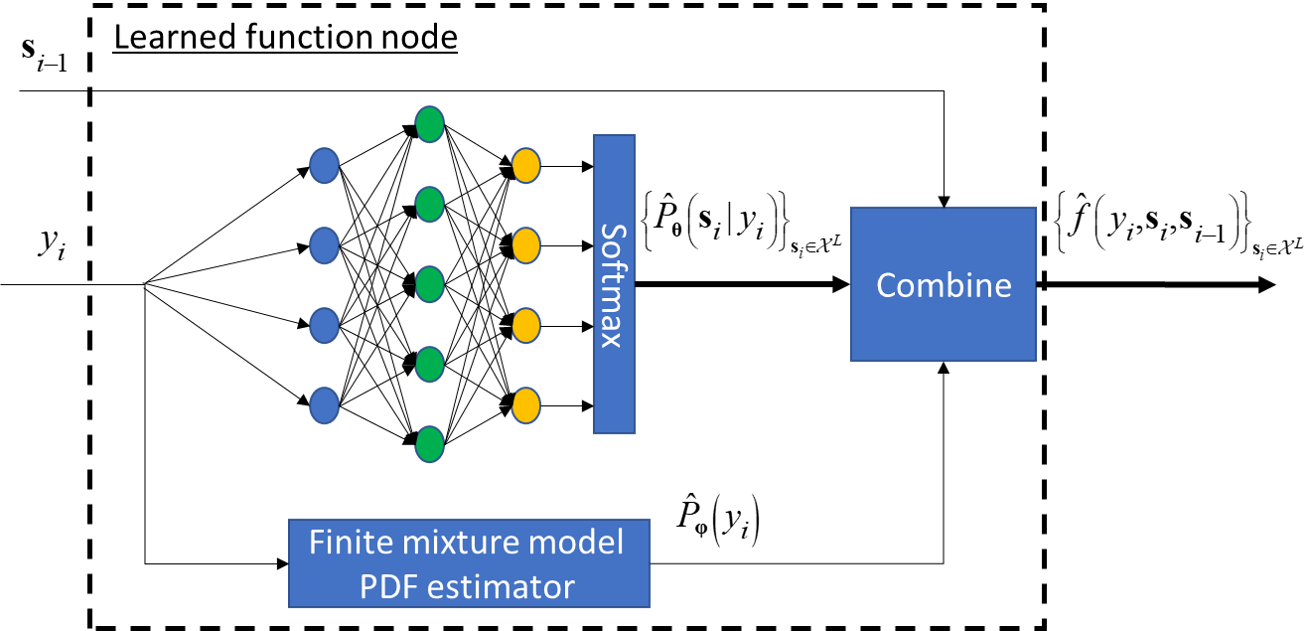}} 
	\caption{Learned function node of BCJRNet illustration.}
	\label{fig:LearnedFunctionNode}	 
\end{figure}

As noted in \cite{shlezinger2019viterbinet}, since $y_i$ is given and may take continuous values while $\myVec{s}$, representing the label, takes discrete values, a natural approach to evaluate $P_{Y|\myVec{S}}(y_i|\myVec{s})$ for  each $\myVec{s}\in\mySet{X}^L$ using \ac{ml} tools is to estimate $P_{\myVec{S}|Y}(\myVec{s}|y_i)$, from which the desired $P_{Y|\myVec{S}}(y_i|\myVec{s})$  can be obtained using Bayes rule via
\begin{equation}
\label{eqn:Bayes}
P_{Y|\myVec{S}}(y_i|\myVec{s}) = |\mySet{X}|^L P_{\myVec{S}|Y}(\myVec{s}|y_i) P_{Y}(y_i).
\end{equation}
In particular, BCJRNet utilizes two parametric models: one for evaluating the conditional  $P_{\myVec{S}|Y}(\myVec{s}|y_i)$, and another for computing the marginal \ac{pdf} $P_{Y}(y_i)$.  A reliable parametric estimate of $P_{\myVec{S}|Y}(\myVec{s}|y_i)$, denoted $P_{\myVec{\theta}}(\myVec{s}|y_i)$,  can be obtained  for each $\myVec{s}\in\mySet{X}^L$ by training a relatively compact classification network with a softmax output layer. For example, in our numerical study in Section \ref{sec:Sims} we use a three-layer network which can be trained with merely $10000$ training samples. In order to estimate the  marginal \ac{pdf} of $Y_i$, we note that since it is given by a stochastic mapping of $\myVec{S}_{i}$, its distribution can be approximated as a mixture model of $|\mySet{X}|^L$ kernel functions \cite{mclachlan2004finite}. Consequently, a parametric estimate of $P_{Y}(y_i)$, denoted $P{\myVec{\varphi}}\left(  y_i\right)$, can be obtained from the training data using mixture density estimation via, e.g., \ac{em}  \cite[Ch. 2]{mclachlan2004finite}, or any other finite mixture model fitting method. The resulting structure  in which the  parameteric  $P_{\myVec{\theta}}(\myVec{s}|y_i)$ and $P{\myVec{\varphi}}\left(  y_i\right)$ are combined into a learned function node using \eqref{eqn:FunctionNode}-\eqref{eqn:Bayes}, is illustrated in Fig. \ref{fig:LearnedFunctionNode}.

\vspace{-0.2cm}
\subsection{Discussion}
\label{subsec:Discussion}
\vspace{-0.1cm}
Learned factor graph combine \ac{ml} and signal processing algorithms, allowing the latter to be carried out in a data-driven manner. Compared to the conventional \ac{ml} approach of utilizing a \ac{dnn} to carry out the desired end-to-end mapping,  \acp{dnn} are utilized here for the simpler task of uncovering the function nodes of a factor graphs. Consequently, networks with a relatively small number of parameters can be utilized, allowing the system to train quickly from small training sets. In particular, in our numerical study detailed in Section \ref{sec:Sims} we used a three layer fully-connected network for learning the factor nodes in BCJRNet, allowing the system to adapt using merely 10000 training bits, e.g., not much more than a typical 4G preamble \cite[Ch. 17]{dahlman20103g}. For comparison, conventional \ac{dnn} architectures used for carrying out the complete symbol detection task in finite-memory channels typically involve deep networks and require much larger training sets, e.g., \cite{farsad2018neural,liao2019deep}. Furthermore, the ability to implement a data-driven receiver which is capable of adapting its mapping using a small data sets paves the way to the possibility of an on-line training receiver, utilizing the inherent structure of coded communications  to train without additional overhead, see, e.g., \cite{shlezinger2019viterbinet} and  \cite{lugosch2018learning}. We leave the exploration of such on-line trained symbol detectors for future work. 

The proposed BCJRNet is one example of how learned factor graphs  give rise to a data-driven implementation of model-based algorithms. We envision learned factor graphs to be a technique for realizing additional systems capable of learning to carry out important methods in communications, signal processing, and statistics, from relatively small labeled data sets. Furthermore, we believe that this concept unifies additional previously proposed systems combining \ac{ml} and signal processing algorithms, such as the hybrid extended Kalman filter proposed in \cite{laufer2018hybrid}. We leave this extension of our study for future work.

\vspace{-0.2cm}
\section{Numerical Evaluations}
\label{sec:Sims}
\vspace{-0.1cm}
In this section we numerically evaluate the performance of BCJRNet, which is obtained by learning the underlying factor graph of a finite-memory channel as detailed in Section~\ref{sec:SymbolDet}, and compare it to the conventional model-based BCJR algorithm. 
In the following, the \ac{dnn} used for computing the conditional distributions in the learned function node in Fig.~\ref{fig:LearnedFunctionNode} consists of three fully-connected layers: a $1\times 100$ layer, a $100\times 50$ layer, and a $50 \times 4$ layer, with intermediate sigmoid and ReLU activations, respectively, and a softmax output layer.  The network is trained using the Adam optimizer  \cite{kingma2014adam} with learning rate of $0.01$. The finite mixture model estimator approximates the distribution as a Gaussian mixture using \ac{em}-based fitting \cite[Ch. 2]{mclachlan2004finite}. The function node is learned from $10000$ training samples, and is tested over $50000$ Monte Carlo simulations. Due to the small number of training samples and the simple  \ac{dnn} architecture, only a few minutes are required to train BCJRNet on a standard CPU.

We simulate two finite-memory channels with memory $L=2$: An \ac{isi} channel with \ac{awgn}, and a Poisson channel. 
To formulate the input-output relationships of these channels, we let  $\{h_1(\gamma),h_2(\gamma)\}$ be $L$ coefficients representing an exponentially decaying profile, given by $h_\tau(\gamma) \triangleq e^{-\gamma(\tau-1)}$ for $\gamma > 0$.
For the \ac{isi} channel with \ac{awgn}, we consider i.i.d. binary phase shift keying inputs, i.e.,  $\mySet{X} = \{-1,1\}$, and the channel output $Y_i$ is related to the input via 
\begin{equation}
\label{eqn:AWGNCh1}
Y_i = \sqrt{\rho} \cdot\sum\limits_{\tau=1}^{L} h_{\tau}(\gamma) X_{i-\tau + 1} + W_i,
\end{equation}
where $W_i$ is a  unit variance \ac{awgn} independent of $X_i$, and $\rho > 0$ represents the \ac{snr}.
For the Poisson channel, the  input represents on-off keying, namely, $\mySet{X} = \{0,1\}$, and the statistical input-output relationship is  
\begin{equation}
\label{eqn:PoissonCh1}
Y_i | \myVec{X} \sim \mathds{P}\left( \sqrt{\rho} \cdot\sum\limits_{\tau=1}^{L}  h_\tau(\gamma)  X_{i-\tau + 1} + 1\right),
\end{equation}
where $\mathds{P}(\lambda)$ is the Poisson distribution with parameter $\lambda > 0$.

BCJRNet is trained for each \ac{snr} value $\rho$, and the \ac{ser} values are averaged over $20$ different channel coefficients $\{h_\tau(\gamma)\}$ with $\gamma$ taking values in $[0.1, 2]$. 
In addition, we evaluate resiliency to inaccurate training by computing the \ac{ser} when the function nodes in Fig. \ref{fig:LearnedFunctionNode} are learned from samples taken from a channel with a noisy estimate of $\{h_\tau(\gamma)\}$. The estimation noise is randomized as an i.i.d. zero-mean Gaussian process with variance $\sigma_e^2$, where we use  $\sigma_e^2 = 0.1$ for the Gaussian channel \eqref{eqn:AWGNCh1}, and $\sigma_e^2 = 0.08$ for the Poisson channel \eqref{eqn:PoissonCh1}. We refer to this scenario as {\em \ac{csi} uncertainty}. Under such uncertainty, the model-based BCJR computes the messages detailed in Example \ref{exm:SP_FSC} using the noisy channel estimate, while BCJRNet is trained using samples taken from different realizations of the noisy $\{h_\tau(\gamma)\}$.  

\begin{figure}
	\centering
	\includegraphics[  width=0.6\columnwidth]{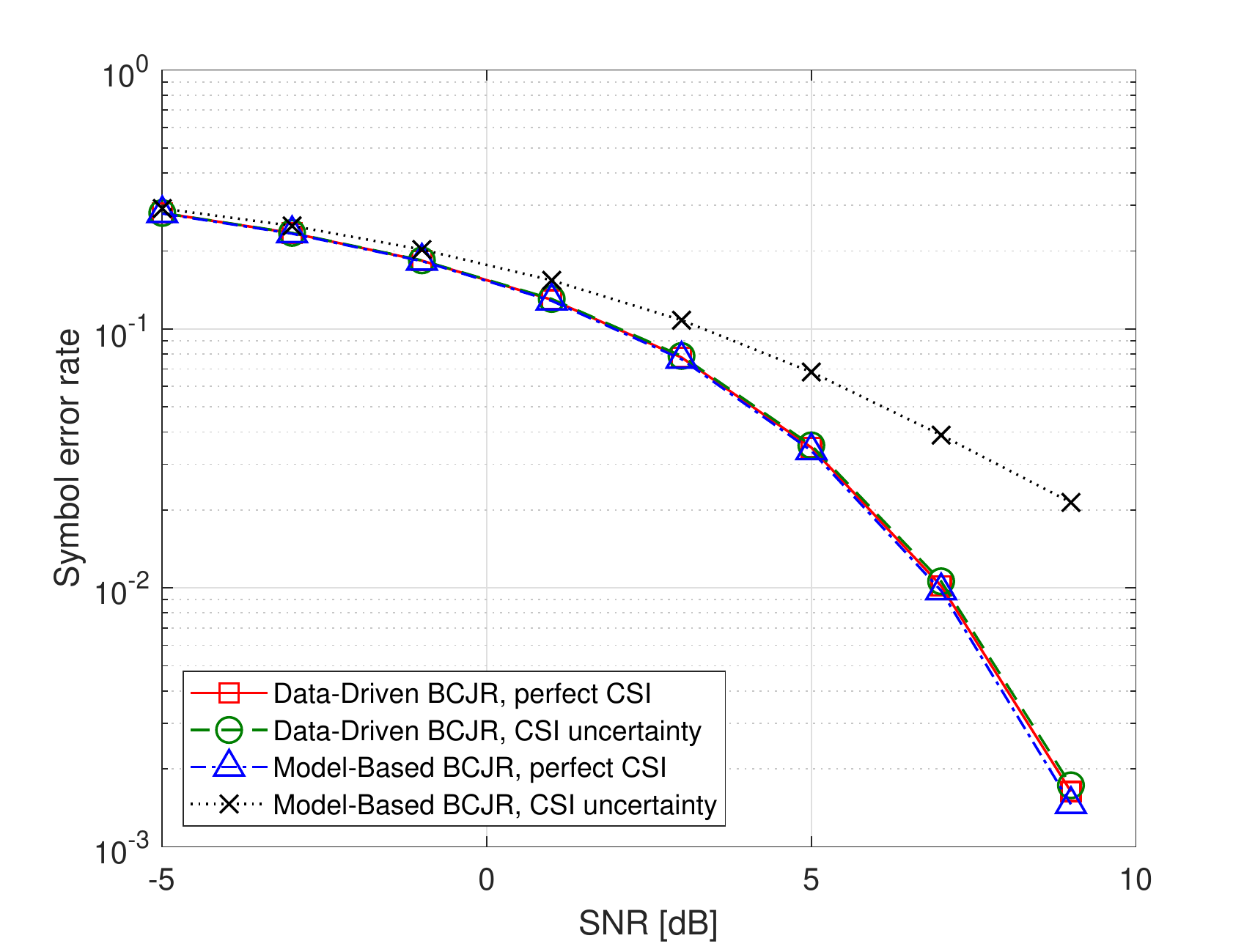}
	\vspace{-0.6cm}
	\caption{\ac{ser} versus \ac{snr}, \ac{isi} channel with \ac{awgn}.
	}
	\label{fig:AWGN}
\end{figure}

\begin{figure}
	\centering
	\includegraphics[  width=0.6\columnwidth]{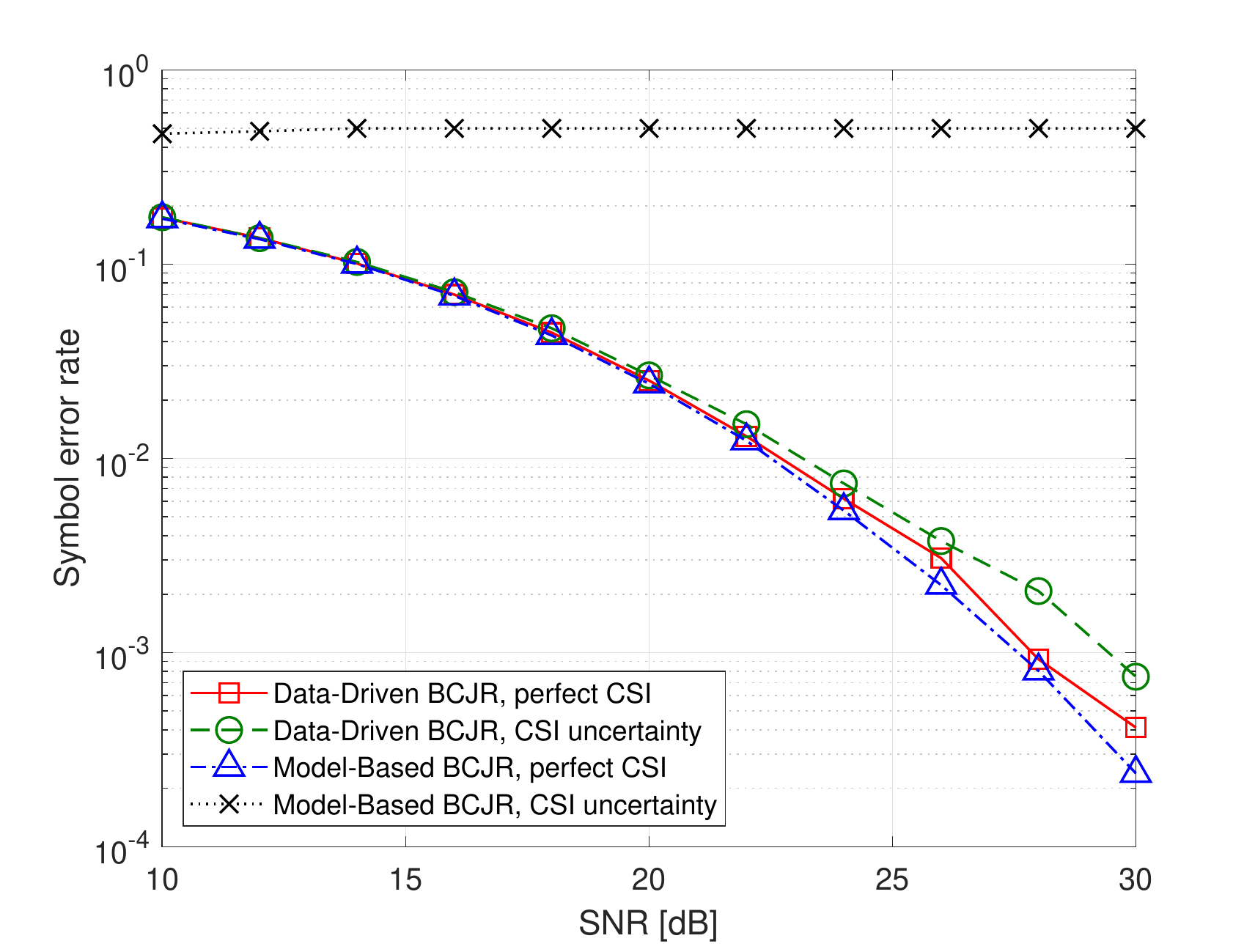}
	\vspace{-0.6cm}
	\caption{\ac{ser} versus \ac{snr}, Poisson channel.
	}
	\label{fig:Poisson}
\end{figure}

The numerically computed \ac{ser} values are depicted in Figs. \ref{fig:AWGN}-\ref{fig:Poisson} for the Gaussian channel and the Poisson channel, respectively. We first note that for both channels, the performance of BCJRNet approaches that of the channel-model-based algorithm, and that the curves effectively coincide for most simulated \ac{snr} values. A small gap is noted for the Poisson channel at high \acp{snr}, which stems from the   model mismatch induced by approximating the marginal \ac{pdf} of $Y_i$ as a Gaussian mixture in the parameteric estimate $P{\myVec{\varphi}}\left(  y_i\right)$. 

We also observe in Figs. \ref{fig:AWGN}-\ref{fig:Poisson}  that the data-driven implementation of the BCJR algorithm is substantially more robust to \ac{csi} uncertainty compared to the channel-model-based detector. For example, in Fig. \ref{fig:AWGN} we note that BCJRNet trained with a variety of different channel conditions achieves roughly the same performance as it does when trained  and tested using samples from the same statistical model. However, the  channel-model-based BCJR algorithm achieves significantly degraded \ac{ser} performance due to imperfect \ac{csi}.  
 
The results presented in this section demonstrate that an \ac{ml}-based receiver can be trained to carry out accurate and robust \ac{map} symbol detection using a relatively small amount of labeled samples. This is achieved by learning the function nodes of the underlying factor graphs, and carrying out the sum-product recursions over the learned graph. 
 
\vspace{-0.2cm}
\section{Conclusions}
\label{sec:Conclusions}
\vspace{-0.1cm} 
We have proposed the concept of data-driven factor graphs methods, which allow a broad family of model-based algorithms to be implemented in a data-driven fashion. Our approach utilizes \ac{ml} tools to adapt the function nodes of the underlying statistical model, allowing to learn the factor graph data, using which the desired scheme can be implemented by message passing over the learned graph. We applied the proposed strategy to learn the factor graph of stationary finite-memory communication channels, yielding BCJRNet, which is a data-driven implementation of the BCJR \ac{map} symbol detector. Our numerical results demonstrate that BCJRNet learns to accurately implement the BCJR algorithm from a relatively small training set, while exhibiting improved robustness to \ac{csi} uncertainty compared to the model-based BCJR detector.
 
\bibliographystyle{IEEEtran}
\bibliography{IEEEabrv,refs}

\begin{thebibliography}{10}
\providecommand{\url}[1]{#1}
\csname url@samestyle\endcsname
\providecommand{\newblock}{\relax}
\providecommand{\bibinfo}[2]{#2}
\providecommand{\BIBentrySTDinterwordspacing}{\spaceskip=0pt\relax}
\providecommand{\BIBentryALTinterwordstretchfactor}{4}
\providecommand{\BIBentryALTinterwordspacing}{\spaceskip=\fontdimen2\font plus
\BIBentryALTinterwordstretchfactor\fontdimen3\font minus
  \fontdimen4\font\relax}
\providecommand{\BIBforeignlanguage}[2]{{%
\expandafter\ifx\csname l@#1\endcsname\relax
\typeout{** WARNING: IEEEtran.bst: No hyphenation pattern has been}%
\typeout{** loaded for the language `#1'. Using the pattern for}%
\typeout{** the default language instead.}%
\else
\language=\csname l@#1\endcsname
\fi
#2}}
\providecommand{\BIBdecl}{\relax}
\BIBdecl

\bibitem{loeliger2007factor}
H.-A. Loeliger, J.~Dauwels, J.~Hu, S.~Korl, L.~Ping, and F.~R. Kschischang,
  ``The factor graph approach to model-based signal processing,''
  \emph{Proceedings of the IEEE}, vol.~95, no.~6, pp. 1295--1322, 2007.

\bibitem{forney2001codes}
G.~D. Forney, ``Codes on graphs: Normal realizations,'' \emph{{IEEE} Trans.
  Inf. Theory}, vol.~47, no.~2, pp. 520--548, 2001.

\bibitem{kschischang2001factor}
F.~R. Kschischang, B.~J. Frey, and H.-A. Loeliger, ``Factor graphs and the
  sum-product algorithm,'' \emph{{IEEE} Trans. Inf. Theory}, vol.~47, no.~2,
  pp. 498--519, 2001.

\bibitem{loeliger2004introduction}
H.-A. Loeliger, ``An introduction to factor graphs,'' \emph{{IEEE} Signal
  Process. Mag.}, vol.~21, no.~1, pp. 28--41, 2004.

\bibitem{bahl1974optimal}
L.~Bahl, J.~Cocke, F.~Jelinek, and J.~Raviv, ``Optimal decoding of linear codes
  for minimizing symbol error rate,'' \emph{{IEEE} Trans. Inf. Theory},
  vol.~20, no.~2, pp. 284--287, 1974.

\bibitem{loeliger2002least}
H.-A. Loeliger, ``Least squares and {Kalman} filtering on {Forney} graphs,'' in
  \emph{Codes, Graphs, and Systems}.\hskip 1em plus 0.5em minus 0.4em\relax
  Springer, 2002, pp. 113--135.

\bibitem{jordan2004graphical}
M.~I. Jordan, ``Graphical models,'' \emph{Statistical Science}, vol.~19, no.~1,
  pp. 140--155, 2004.

\bibitem{colavolpe2005application}
G.~Colavolpe and G.~Germi, ``On the application of factor graphs and the
  sum-product algorithm to {ISI} channels,'' \emph{{IEEE} Trans. Commun.},
  vol.~53, no.~5, pp. 818--825, 2005.

\bibitem{drost2007factor}
R.~J. Drost and A.~C. Singer, ``Factor-graph algorithms for equalization,''
  \emph{{IEEE} Trans. Signal Process.}, vol.~55, no.~5, pp. 2052--2065, 2007.

\bibitem{som2011low}
P.~Som, T.~Datta, N.~Srinidhi, A.~Chockalingam, and B.~S. Rajan,
  ``Low-complexity detection in large-dimension mimo-isi channels using
  graphical models,'' \emph{{IEEE} J. Sel. Topics Signal Process.}, vol.~5,
  no.~8, pp. 1497--1511, 2011.

\bibitem{niu2005factor}
H.~Niu, M.~Shen, J.~A. Ritcey, and H.~Liu, ``A factor graph approach to
  iterative channel estimation and ldpc decoding over fading channels,''
  \emph{{IEEE} Trans. Wireless Commun.}, vol.~4, no.~4, pp. 1345--1350, 2005.

\bibitem{lehmann2016factor}
F.~Lehmann, ``A factor graph approach to iterative channel estimation,
  detection, and decoding for two-path successive relay networks,''
  \emph{{IEEE} Trans. Wireless Commun.}, vol.~15, no.~8, pp. 5414--5429, 2016.

\bibitem{shlezinger2019viterbinet}
N.~Shlezinger, N.~Farsad, Y.~C. Eldar, and A.~J. Goldsmith, ``{ViterbiNet}: A
  deep learning based {Viterbi} algorithm for symbol detection,'' \emph{IEEE
  Trans. Wireless Commun., early access}, 2019.

\bibitem{shlezinger2019deepSIC}
N.~Shlezinger, R.~Fu, and Y.~C. Eldar, ``{DeepSIC}: Deep soft interference
  cancellation for multiuser {MIMO} detection,'' \emph{Submitted to IEEE J.
  Sel. A. Inform. Theory}, 2019.

\bibitem{forney1973viterbi}
G.~D. Forney, ``The {Viterbi} algorithm,'' \emph{Proceedings of the IEEE},
  vol.~61, no.~3, pp. 268--278, 1973.

\bibitem{choi2000iterative}
W.-J. Choi, K.-W. Cheong, and J.~M. Cioffi, ``Iterative soft interference
  cancellation for multiple antenna systems.'' in \emph{Proc. WCNC}, 2000, pp.
  304--309.

\bibitem{lecun2015deep}
Y.~LeCun, Y.~Bengio, and G.~Hinton, ``Deep learning,'' \emph{nature}, vol. 521,
  no. 7553, p. 436, 2015.

\bibitem{weiss2001optimality}
Y.~Weiss and W.~T. Freeman, ``On the optimality of solutions of the max-product
  belief-propagation algorithm in arbitrary graphs,'' \emph{{IEEE} Trans. Inf.
  Theory}, vol.~47, no.~2, pp. 736--744, 2001.

\bibitem{pearl1986fusion}
J.~Pearl, ``Fusion, propagation, and structuring in belief networks,''
  \emph{Artificial intelligence}, vol.~29, no.~3, pp. 241--288, 1986.

\bibitem{wiberg1995codes}
N.~Wiberg, H.-A. Loeliger, and R.~Kotter, ``Codes and iterative decoding on
  general graphs,'' \emph{European Transactions on telecommunications}, vol.~6,
  no.~5, pp. 513--525, 1995.

\bibitem{bengio2009learning}
Y.~Bengio, ``Learning deep architectures for {AI},'' \emph{Foundations and
  trends{\textregistered} in Machine Learning}, vol.~2, no.~1, pp. 1--127,
  2009.

\bibitem{cioffi2008equalization}
J.~M. Cioffi, ``Sequence detection,'' \emph{EE379B Course notes chapter 9.
  Stanford University}, 2008.

\bibitem{mclachlan2004finite}
G.~McLachlan and D.~Peel, \emph{Finite mixture models}.\hskip 1em plus 0.5em
  minus 0.4em\relax John Wiley \& Sons, 2004.

\bibitem{dahlman20103g}
E.~Dahlman, S.~Parkvall, J.~Skold, and P.~Beming, \emph{{3G} evolution: {HSPA}
  and {LTE} for mobile broadband}.\hskip 1em plus 0.5em minus 0.4em\relax
  Academic press, 2010.

\bibitem{farsad2018neural}
N.~Farsad and A.~Goldsmith, ``Neural network detection of data sequences in
  communication systems,'' \emph{{IEEE} Trans. Signal Process.}, vol.~66,
  no.~21, pp. 5663--5678, 2018.

\bibitem{liao2019deep}
Y.~Liao, N.~Farsad, N.~Shlezinger, Y.~C. Eldar, and A.~J. Goldsmith, ``Deep
  neural network symbol detection for millimeter wave communications,''
  \emph{arXiv preprint arXiv:1907.11294}, 2019.

\bibitem{lugosch2018learning}
L.~Lugosch and W.~J. Gross, ``Learning from the syndrome,'' in \emph{2018 52nd
  Asilomar Conference on Signals, Systems, and Computers}.\hskip 1em plus 0.5em
  minus 0.4em\relax IEEE, 2018, pp. 594--598.

\bibitem{laufer2018hybrid}
B.~Laufer-Goldshtein, R.~Talmon, and S.~Gannot, ``A hybrid approach for speaker
  tracking based on tdoa and data-driven models,'' \emph{{IEEE/ACM} Trans.
  Audio, Speech, Language Process.}, vol.~26, no.~4, pp. 725--735, 2018.

\bibitem{kingma2014adam}
D.~P. Kingma and J.~Ba, ``Adam: A method for stochastic optimization,''
  \emph{arXiv preprint arXiv:1412.6980}, 2014.

\end{thebibliography}

\end{document}